\def\year{2021}\relax
\documentclass[letterpaper]{article} 
\usepackage{aaai21}  
\usepackage{times}  
\usepackage{helvet} 
\usepackage{courier}  
\usepackage[hyphens]{url}  
\usepackage{graphicx} 
\usepackage{natbib}
\usepackage{subfigure}
\usepackage{graphicx}
\usepackage{multirow}
\usepackage{algorithmic}
\usepackage{textcomp}
\usepackage{xcolor}
\usepackage{colortbl,booktabs}
\usepackage[ruled,vlined]{algorithm2e} 
\usepackage{times}  
\usepackage{soul}
\usepackage{bm}
\usepackage{changepage}

\usepackage{amsthm}
\usepackage{mathrsfs}
\usepackage{amssymb}
\usepackage{amsmath}
\newtheorem{theorem}{\textbf{Theorem}}
\newtheorem{theorem*}{Theorem}

\newtheorem{Definition}{Definition}

\newtheorem{definition*}{Problem}

\DeclareMathOperator*{\argmin}{argmin}
\DeclareMathOperator*{\argmax}{argmax}

\usepackage{caption} 
\title{How does the Combined Risk Affect the Performance of Unsupervised Domain Adaptation Approaches?}
\author{
Li Zhong$^{1,2, }$\footnote{Equal Contribution. Work done at AAII, UTS.},
Zhen Fang$^{2,*}$,
Feng Liu$^{2,*}$,
Jie Lu$^{2, }$\footnote{Corresponding Author},
Bo Yuan$^1$,
Guangquan Zhang$^2$\\}

\affiliations{
$^1$Shenzhen International Graduate School, Tsinghua University\\
$^2$Centre for Artificial Intelligence, University of Technology Sydney\\
zhongl18@mails.tsinghua.edu.cn, 
feng.liu@uts.edu.au,
zhen.fang@student.uts.edu.au,\\ yuanb@sz.tsinghua.edu.cn,
\{guangquan.zhang, jie.lu\}@uts.edu.au
}

\begin{document}
\maketitle
\begin{abstract}
\textit{Unsupervised domain adaptation} (UDA) aims to train a target classifier with labeled samples from the source domain and unlabeled samples from the target domain. Classical UDA learning bounds show that target risk is upper bounded by three terms: source risk, distribution discrepancy, and combined risk. Based on the assumption that the combined risk is a small fixed value, methods based on this bound train a target classifier by only minimizing estimators of the source risk and the distribution discrepancy. However, the combined risk may increase when minimizing both estimators, which makes the target risk uncontrollable.
Hence the target classifier cannot achieve ideal performance if we fail to control the combined risk.
To control the combined risk, the key challenge takes root in the unavailability of the labeled samples in the target domain.
To address this key challenge, we propose a method named E-MixNet. E-MixNet employs \textit{enhanced mixup}, a generic vicinal distribution, on the labeled source samples and pseudo-labeled target samples to calculate a proxy of the combined risk. Experiments show that the proxy can effectively curb the increase of the combined risk when minimizing the source risk and distribution discrepancy. Furthermore, we show that if the proxy of the combined risk is added into loss functions of four representative UDA methods, their performance is also improved.
\end{abstract}

\section{Introduction}
\begin{figure}[ht]
	\centering
	\includegraphics[scale=0.38,trim=10 15 0 0, clip]{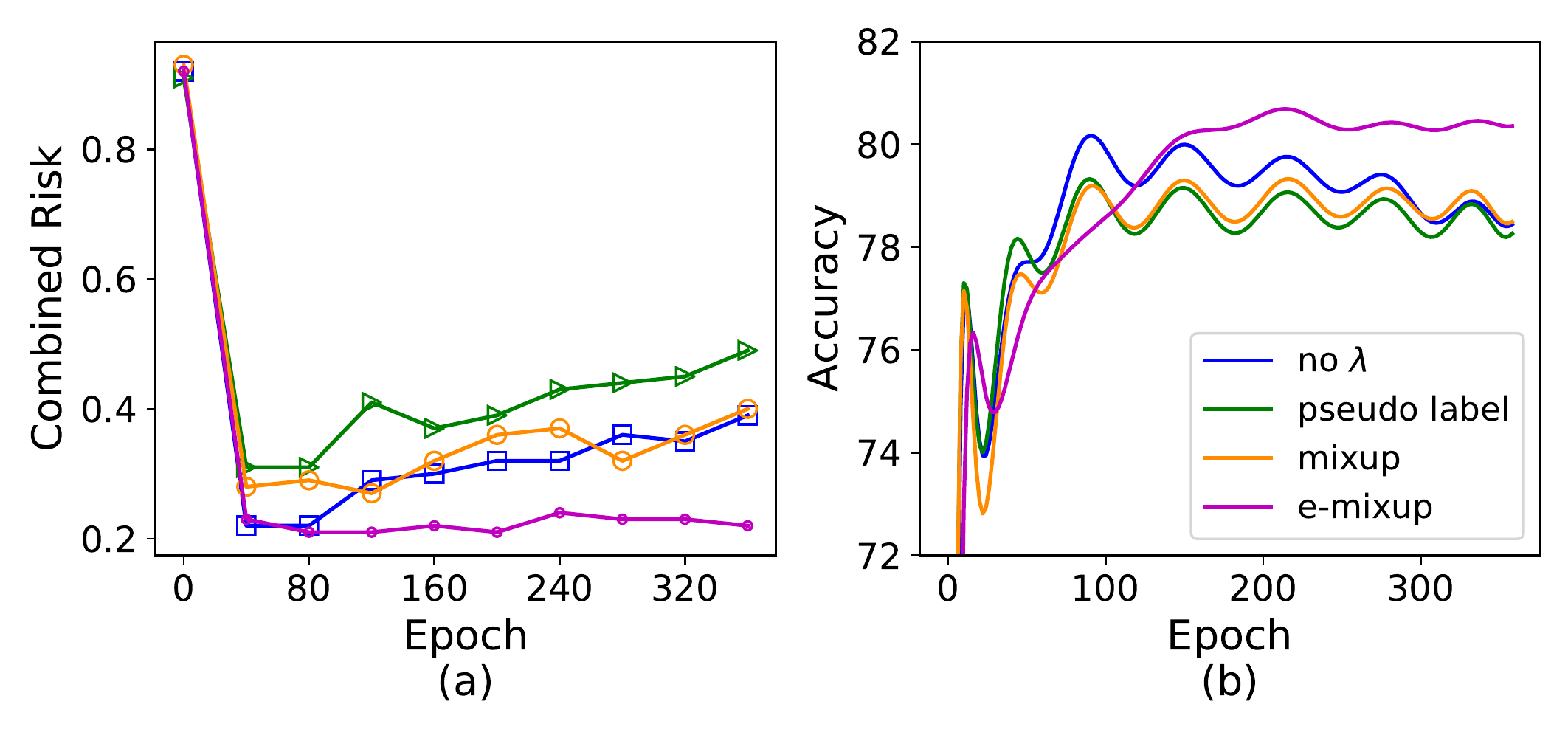}
	\caption{The values of combined risk and accuracy on the task C $\rightarrow$ P on {Image-CLEF}. The left figure shows the value of combined risk. The right figure shows the accuracy of the task. Blue line: ignore the optimization of combined risk. Green line: optimize the combined risk by source samples and target samples with high confidence. Orange line: optimize combined risk by the proxy formulated by mixup. Purple line: optimize the proxy formulated by e-mixup.}
	\label{fig:combined_risk}
	\vspace{-1em}
\end{figure}

\emph{Domain Adaptation} (DA) aims to train a target-domain classifier with samples from source and target domains \cite{lu2015transfer}. When the labels of samples in the target domain are unavailable, DA is known as \emph{unsupervised DA} (UDA) \cite{zhong2020bridging,fang2020open}, which has been applied to address diverse real-world problems, such as computer version \cite{zhang2020clarinet,Semantic_Transferable_Dong_ICCV2019,Weakly_Supervised_Dong_TCSVT2020}, natural language processing \cite{lee2019zero, guo2020multi}, and recommender system \cite{zhang2017cross,yu2019input,lu2020dual} 



Significant theoretical advances have been achieved in UDA. Pioneering theoretical work was proposed by Ben-David et al. \shortcite{ben2007analysis}. This work shows that the target risk is upper bounded by three terms: source risk, marginal distribution discrepancy, and  \textit{combined risk}. This earliest learning bound has been extended from many perspectives, such as considering more surrogate loss functions \cite{pmlr-v97-zhang19i} or  distributional discrepancies \cite{mohri2012new, shen2018wasserstein} (see \cite{redko2020survey} as a survey).
Recently, Zhang et al. \shortcite{pmlr-v97-zhang19i} proposed a new distributional discrepancy termed Margin Disparity Discrepancy and developed a tighter and more practical UDA learning bound.

The UDA learning bounds proposed by \cite{ben2007analysis, ben2010theory} and the recent UDA learning bounds proposed by \cite{shen2018wasserstein, xu2019adversarial, zhang2020unsupervised} consist of three terms: source risk, marginal distribution discrepancy, and combined risk. Minimizing the source risk aims to obtain a source-domain classifier, and minimizing the distribution discrepancy aims to learn domain-invariant features so that the source-domain classifier can perform well on the target domain. The combined risk embodies the \textit{adaptability} between the source and target domains \cite{ben2010theory}. In particularly, when the hypothesis space is fixed, the combined risk is a constant.

Based on the UDA learning bounds where the combined risk is assumed to a small constant, many existing UDA methods focus on learning domain-invariant features \cite{fang2019unsupervised,What_Transferred_Dong_CVPR2020,CSCL_Dong_ECCV2020,liu2019butterfly} by minimizing the estimators of the source risk and the distribution discrepancy. In the learned feature space, the source and target distributions are similar while the source-domain classifier is required to achieve a small error. Furthermore, the generalization error of the source-domain classifier is expected to be small in the target domain.

However, the combined risk may increase when learning the domain-invariant features, and the increase of the combined risk may degrade the performance of the source-domain classifier in the target domain. As shown in Figure~\ref{fig:combined_risk}, we calculate the value of the combined risk and accuracy on a real-world UDA task (see the green line). The performance worsens with the increased combined risk.  Zhao et al. \shortcite{Zhao2019OnLI} also pointed out the increase of combined risk causes the failure of source-domain classifier on the target domain.

To investigate how the combined risk affect the performance on the domain-invariant features, we rethink and develop the UDA learning bounds by introducing  feature transformations. In the new bound (see Eq. \eqref{eq:bound_dl}), the combined risk is a function related to feature transformation but not a constant (compared to bounds in \cite{ben2010theory}).
We also reveal that the combined risk is deeply related to the conditional distribution discrepancy (see {Theorem 3}). {Theorem 3} shows that, the conditional distribution discrepancy will increase when the combined risk increases. Hence, it is hard to achieve satisfied target-domain accuracy if we only focus on learning domain-invariant features and omit to control the combined risk.

To estimate the combined risk, the key challenge  takes root in the \emph{unavailability}  of the labeled  samples in the target domain. 
A simple solution is to leverage the pseudo labels with high confidence in the target domain to estimate the combined risk. However, since samples with high confidence are insufficient, the value of the combined risk may still increase (see the green line in Figure~\ref{fig:combined_risk}).
Inspired by semi-supervised learning methods, an advanced solution is to directly use the \textit{mixup} technique to augment pseudo-labeled target samples, which can slightly help us estimate the combined risk better than the simple solution (see the orange line in Figure~\ref{fig:combined_risk}).


However, the target-domain pseudo labels provided by the source-domain classifier may be inaccurate due to the discrepancy between domains, which causes that \textit{mixup} may not perform well with inaccurate labels. To mitigate the issue, we propose \textit{enhanced mixup} (e-mixup) to substitute \textit{mixup} to compute a proxy of the combined risk. The purple line in Figure~\ref{fig:combined_risk} shows that the proxy based on e-mixup can significantly boost the performance. 
Details of the proxy is shown in section Motivation. 

To the end, we design a novel UDA method referred to E-MixNet. E-MixNet learns the target-domain classifier by simultaneously minimizing the source risk, the marginal distribution discrepancy, and the proxy of combined risk. Via minimizing the proxy of combined risk, we control the increase of combined risk effectively, thus, control the conditional distribution discrepancy between two domains.

We conduct experiments on three public datasets (Office-31, Office-Home, and Image-CLEF) and compare E-MixNet with a series of existing state-of-the-art methods. Furthermore, we introduce the proxy of the combined risk into four representative UDA methods (\textit{i.e.}, DAN \cite{long2015learning}, DANN \cite{ganin2016domain}, CDAN \cite{long2018conditional}, SymNets \cite{zhang2019domain}). Experiments show that E-MixNet can outperform all baselines, and the four representative methods can achieve better performance if the proxy of the combined risk is added into their loss functions.

\section{Problem Setting and Concepts}
In this section, we introduce the definition of UDA, then introduce some important concepts used in this paper. 

Let $\mathcal{X}\subset \mathbb{R}^d$ be a feature space and $\mathcal{Y}:=\{{\mathbf{y}}_c\}_{c=1}^K$ be the label space, where the label ${\mathbf{y}}_c\in \mathbb{R}^{{K}}$ is a one-hot vector, whose $c$-th coordinate is $1$ and the other coordinates are $0$. 
\begin{Definition}[Domains in UDA]\label{d3}Given random variables $X_s, X_t \in \mathcal{X}$, $Y_s,  {Y}_t \in \mathcal{Y}$, the {source} and {target domains} are joint distributions $P_{X_s {Y}_s}$ and $P_{X_t {Y}_t}$ with $P_{X_sY_s}\neq P_{X_t{Y}_t}$.
\end{Definition}
Then, we propose the UDA problem as follows.
\begin{definition*}[UDA]
Given independent and identically distributed (\textit{i.i.d.})~labeled samples ${D}_s=\{(\mathbf{x}_s^i,\mathbf{y}_s^i)\}^{{n}_s}_{i=1}$ drawn from the source domain $P_{X_s{Y}_s}$ and \textit{i.i.d.}~unlabeled samples $D_t=\{\mathbf{x}_t^i\}^{n_t}_{i=1}$ drawn from the target marginal distribution $P_{X_t}$, {the aim} of UDA is to train a classifier  $f:\mathcal{X}\rightarrow \mathcal{Y}$ with ${D}_s$ and $D_t$ such that
$f$ classifies accurately target data $D_t$.
\end{definition*}

Given a loss function $\ell: \mathbb{R}^K\times   \mathbb{R}^K\rightarrow \mathbb{R}_{\geq 0}$ and any scoring functions $\mathbf{C},\mathbf{C}^{\prime}$ from function space $\{\mathbf{C}:\mathcal{X}\rightarrow \mathbb{R}^K\}$, source risk, target risk and classifier discrepancy are
\begin{equation*}
\begin{split}
   & R_s^{\ell}(\mathbf{C}):=\mathbb{E}_{(\mathbf{x},\mathbf{y})\sim P_{X_sY_s}} \ell(\mathbf{C}(\mathbf{x}),\mathbf{y}), \\ &R_t^{\ell}(\mathbf{C}):=\mathbb{E}_{(\mathbf{x},\mathbf{y})\sim P_{X_tY_t}} \ell(\mathbf{C}(\mathbf{x}),\mathbf{y}),\\
   & R_s^{\ell}(\mathbf{C}^\prime,\mathbf{C}):=\mathbb{E}_{\mathbf{x}\sim P_{X_s}} \ell(\mathbf{C}^\prime(\mathbf{x}),\mathbf{C}(\mathbf{x})), \\ &R_t^{\ell}(\mathbf{C}^\prime,\mathbf{C}):=\mathbb{E}_{\mathbf{x}\sim P_{X_t}} \ell(\mathbf{C}^\prime(\mathbf{x}),\mathbf{C}(\mathbf{x})).
    \end{split}
\end{equation*}

Lastly, we define the disparity discrepancy based on double losses, which will be used to design our method.
\begin{Definition}[Double Loss Disparity Discrepancy]\label{doubleloss}Given distributions $P, Q$ over some feature space $\widetilde{\mathcal{X}}$, two losses $\ell_s,\ell_t$,  a hypothesis space $\mathcal{H}\subset \{\mathbf{C}:\widetilde{\mathcal{X}}\rightarrow \mathbb{R}^K\}$ and any scoring function $\mathbf{C}\in \mathcal{H}$, then the double loss disparity discrepancy $d_{\mathbf{C}, \mathcal{H}}^{\ell_s\ell_t}(P,Q)$ is
\begin{equation}
\label{eq:dbl_dd}
\begin{split}
 \sup_{\mathbf{C}^\prime \in \mathcal{H}} \big (R_P^{\ell_t}(\mathbf{C}^\prime,\mathbf{C})- R_Q^{\ell_s}(\mathbf{C}^\prime,\mathbf{C}) \big ),
 \end{split}
\end{equation}
where
\begin{equation*}
\begin{split}
 & R_P^{\ell_t}(\mathbf{C}^\prime,\mathbf{C}):=\mathbb{E}_{\mathbf{x}\sim P} \ell_t(\mathbf{C}^\prime(\mathbf{x}),\mathbf{C}(\mathbf{x})),
 \\ & R_Q^{\ell_s}(\mathbf{C}^\prime,\mathbf{C}):=\mathbb{E}_{\mathbf{x}\sim Q} \ell_s(\mathbf{C}^\prime(\mathbf{x}),\mathbf{C}(\mathbf{x})).
  \end{split}
\end{equation*}
\end{Definition}
When losses $\ell_s,\ell_t$ are the margin loss \cite{pmlr-v97-zhang19i}, the double loss disparity discrepancy is known as the  Margin Disparity Discrepancy \cite{pmlr-v97-zhang19i}.

Compared with the classical discrepancy distance \cite{yishay2009domain}:
\begin{equation}
\label{eq:dd}
\begin{split}
    &d^{\ell}_{\mathcal{H}}(P,Q):=  \sup_{\mathbf{C}^\prime,\mathbf{C} \in \mathcal{H}} \big | R_P^{\ell}(\mathbf{C}^\prime,\mathbf{C})- R_Q^{\ell}(\mathbf{C}^\prime,\mathbf{C}) \big |,
    \end{split}
\end{equation}
double loss disparity discrepancy is tighter and more flexible.

\begin{theorem}[{DA Learning Bound}]\label{classicalbound}
Given a loss $\ell$ satisfying the triangle inequality and a hypothesis space $\mathcal{H}\subset \{\mathbf{C}:\mathcal{X}\rightarrow \mathbb{R}^K \}$, then for any $\mathbf{C}\in \mathcal{H}$, we have
\begin{equation*}
    R_t^{\ell}(\mathbf{C})\leq R_s^{\ell}(\mathbf{C})+d^{\ell}_{\mathcal{H}}(P_{X_s},P_{X_t})+\lambda^{\ell},
\end{equation*}
where $d^{\ell}_{\mathcal{H}}$ is the discrepancy distance defined in Eq. \eqref{eq:dd} and $\lambda^{\ell}:=\min_{\mathbf{C}^{*}\in \mathcal{H}} \big ( R_s^{\ell}(\mathbf{C}^*)+R_t^{\ell}(\mathbf{C}^*)\big )$ known as the combined risk.
\end{theorem}
In Theorem \ref{classicalbound}, when the hypothesis space $\mathcal{H}$ and the loss $\ell$ are fixed, the combined risk is a fixed constant. Note that, under certain assumptions, the target risk can be upper bounded only by the first two terms \cite{gong2016domain,gong2020domain}, which is also a promising research direction. 

\section{Theoretical Analysis}
Here we introduce our main theoretical results. All proofs can be found at https://github.com/zhonglii/E-MixNet.

\subsection{Rethinking DA Learning Bound}
Many existing UDA methods \cite{wang2019unsupervised, zou2019consensus, tang2020discriminative} learn a suitable feature transformation $\mathbf{G}$ such  that  the discrepancy between transformed distributions $P_{\mathbf{G}(X_s)}$ and $P_{\mathbf{G}(X_t)}$ is reduced. By introducing the transformation $\mathbf{G}$ in the classical DA learning bound, we discover that the combined error $\lambda^{\ell}$ is not a fixed value.
\begin{theorem}\label{theorem2}
Given a loss $\ell$ satisfying the triangle inequality, a transformation space $\mathcal{G}\subset \{\mathbf{G}:\mathcal{X}\rightarrow \mathcal{X}_{\rm new}\}$ and  a hypothesis space $\mathcal{H}\subset \{\mathbf{C}:\mathcal{X}_{\rm new}\rightarrow \mathbb{R}^K \}$, then for any $\mathbf{G}\in \mathcal{G}$ and $\mathbf{C}\in \mathcal{H}$,
\begin{equation*}
    R_t^{\ell}(\mathbf{C}\circ \mathbf{G})\leq R_s^{\ell}(\mathbf{C}\circ \mathbf{G})+d^{\ell}_{\mathcal{H}}(P_{\mathbf{G}(X_s)},P_{\mathbf{G}(X_t)})+\lambda^{\ell}(\mathbf{G}),
\end{equation*}
where $d^{\ell}_{\mathcal{H}}$ is the discrepancy distance defined in Eq. \eqref{eq:dd} and
\begin{equation}\label{combined risk}
\lambda^{\ell}(\mathbf{G}):=\min_{\mathbf{C}^*\in \mathcal{H}} \big ( R_s^{\ell}(\mathbf{C}^*\circ \mathbf{G})+R_t^{\ell}(\mathbf{C}^*\circ \mathbf{G})\big )
\end{equation}
known as the combined risk.
\end{theorem}

According to Theorem \ref{theorem2}, it is not enough to minimize the source risk and distribution discrepancy by seeking the optimal classifier $\mathbf{C}$  and optimal transformation $\mathbf{G}$ from spaces $\mathcal{H}$ and $\mathcal{G}$. Because we cannot guarantee the value of combined risk $\lambda^{\ell}(\mathbf{G})$ is always small during the training process.

For convenience, we define
\begin{equation}\label{combined}
    \Lambda^{\ell}(\mathbf{C},\mathbf{G}):=R_s^{\ell}(\mathbf{C}\circ \mathbf{G})+R_t^{\ell}(\mathbf{C}\circ \mathbf{G}),
\end{equation}
hence, $\lambda^{\ell}(\mathbf{G})=\min_{\mathbf{C}^* \in \mathcal{H}} \Lambda^{\ell}(\mathbf{C}^*,\mathbf{G})$.
\subsection{Meaning of Combined Risk $\lambda^{\ell}(\mathbf{G})$}
To future understand the meaning of the combined risk $\lambda^{\ell}(\mathbf{G})$, we prove the following Theorem.
\begin{theorem}
\label{theorem3}
Given a symmetric loss $\ell$ satisfying the triangle inequality, a feature transformation $\mathbf{G} \in \mathcal{G}\subset\{\mathbf{G}:\mathcal{X} \rightarrow \mathcal{X}_{\rm new}\}$, a hypothesis space $\mathcal{H} \subset\{\mathbf{C}: \mathcal{X}_{\rm new}\rightarrow\mathbb{R}^K\}$, and
\begin{equation*}
\mathbf{C}_s= \argmin_{\mathbf{C}\in\mathcal{H}}R_s^{\ell}(\mathbf{C}\circ\mathbf{G}),~~ 
\mathbf{C}_t = \argmin_{\mathbf{C}\in\mathcal{H}}R_t^{\ell}(\mathbf{C}\circ\mathbf{G}),
\end{equation*}
then
\begin{equation*}
\begin{split}
2\lambda^{\ell}(\mathbf{G})-\delta&\leq R_s^{\ell}(\mathbf{C}_t\circ\mathbf{G}) + R_t^{\ell}(\mathbf{C}_s\circ\mathbf{G}) \\&\leq 2\lambda^{\ell}(\mathbf{G}) + d^{\ell}_{\mathcal{H}}(P_{\mathbf{G}(X_s)}, P_{\mathbf{G}(X_t)}) + \delta,
\end{split}
\end{equation*}
where $\lambda^{\ell}(\mathbf{G})$ is defined in Eq. (\ref{combined risk}), $\delta := R_s^{\ell}(\mathbf{C}_s\circ\mathbf{G}) + R_t^{\ell}(\mathbf{C}_t\circ\mathbf{G})$ known as the approximation error and $d^{\ell}_{\mathcal{H}}$ is the discrepancy distance defined in Eq. \eqref{eq:dd}.
\end{theorem}

Theorem 3 implies that the combined risk $\lambda^{\ell}(\mathbf{G})$ is deeply related to the optimal classifier discrepancy
\begin{equation*}
    R_s^{\ell}(\mathbf{C}_t\circ\mathbf{G}) + R_t^{\ell}(\mathbf{C}_s\circ\mathbf{G}),
\end{equation*}
which can be regarded as the conditional distribution discrepancy between $P_{Y_s|\mathbf{G}(X_s)}$ and $P_{Y_t|\mathbf{G}(X_t)}$. If $\lambda^{\ell}(\mathbf{G})$ increases, the conditional distribution discrepancy is larger.

\subsection{Double Loss DA Learning Bound}
Note that there exist methods, such as MDD \cite{pmlr-v97-zhang19i}, whose source and target losses are different. To understand these UDA methods and bridge the gap between theory and algorithms, we develop the classical   DA learning bound to a more general scenario.
\begin{theorem}\label{theorem4}
Given losses $\ell_s$ and $\ell_t$ satisfying the triangle inequality, a transformation space $\mathcal{G}\subset \{\mathbf{G}:\mathcal{X}\rightarrow \mathcal{X}_{\rm new}\}$ and  a hypothesis space $\mathcal{H}\subset \{\mathbf{C}:\mathcal{X}_{\rm new}\rightarrow \mathbb{R}^K \}$, then for any $\mathbf{G}\in \mathcal{G}$ and $\mathbf{C}\in \mathcal{H}$, then $R_t^{\ell_t}(\mathbf{C}\circ \mathbf{G})$ is  bounded by
\begin{equation}
\label{eq:bound_dl}
\begin{split}
     R_s^{\ell_s}(\mathbf{C}\circ \mathbf{G})+d^{\ell_s\ell_t}_{\mathbf{C},\mathcal{H}}(P_{\mathbf{G}(X_s)},P_{\mathbf{G}(X_t)}) +\lambda^{\ell_s\ell_t}(\mathbf{G}),
    \end{split}
\end{equation}
where $d^{\ell_s\ell_t}_{\mathbf{C},\mathcal{H}}$ is the double loss disparity discrepancy defined in Eq. \eqref{eq:dbl_dd}
and $\lambda^{\ell_s\ell_t}(\mathbf{G})$ is the combined risk:
\begin{equation}\label{combined risk2}
\lambda^{\ell_s\ell_t}(\mathbf{G}):=\min_{\mathbf{C}^*\in \mathcal{H}} \Lambda^{\ell_1 \ell_2}(\mathbf{C}^*,\mathbf{G}),~~~{\rm here}
\end{equation}
\begin{equation}\label{DoubleCombinedRisk}
\begin{split}
 \Lambda^{\ell_1 \ell_2}(\mathbf{C}^*,\mathbf{G}):=R^{\ell_s}_s(\mathbf{C}^*\circ \mathbf{G})+R^{\ell_t}_t(\mathbf{C}^*\circ \mathbf{G}).
\end{split}
\end{equation}
\end{theorem}

In Theorem 4, the condition that $\ell_s$ and $\ell_t$ satisfy the triangle inequality, can be replaced by a weaker condition:
\begin{equation*}
\begin{split}
 & R_t^{\ell_t}(\mathbf{C}\circ \mathbf{G})\leq R_t^{\ell_t}(\mathbf{C}^\prime\circ \mathbf{G},\mathbf{C}\circ \mathbf{G})+R_t^{\ell_t}(\mathbf{C}^\prime\circ \mathbf{G}), 
\\ &
R_s^{\ell_s}(\mathbf{C}^\prime\circ \mathbf{G},\mathbf{C}\circ \mathbf{G})\leq R_s^{\ell_s}(\mathbf{C}^\prime\circ \mathbf{G})+R_s^{\ell_s}(\mathbf{C}\circ \mathbf{G}).
\end{split}
\end{equation*}
If we set $\ell_s,\ell_t$ as the margin loss, $\ell_s,\ell_t$ do not satisfy the triangle inequality but they satisfy above condition.
\section{Proposed Method: E-MixNet}
Here we introduce motivation and details of our method.
\subsection{Motivation}

Theorem 3 has shown that the combined risk is related to the conditional distribution discrepancy. As the increase of the combined risk, the conditional distribution discrepancy is increased. 
Hence, omitting the importance of the combined risk may make negative impacts on the target-domain accuracy. Figure \ref{fig:combined_risk} (blue line) verifies our observation.

To control the combined risk, we consider the following problem.
\begin{equation}\label{optim1}
\begin{split}
   & \min_{\mathbf{G}\in \mathcal{G}} \lambda^{\ell}(\mathbf{G})=\min_{\mathbf{G}\in \mathcal{G},\mathbf{C}^*\in \mathcal{H}} \Lambda^{\ell}(\mathbf{C}^*,\mathbf{G}),
    \end{split}
\end{equation}
where $\Lambda^{\ell}(\mathbf{C}^*,\mathbf{G})$ is defined in Eq. (\ref{combined}). Eq. \eqref{optim1} shows we can control the combined risk by minimizing $\Lambda^{\ell}(\mathbf{C}^*,\mathbf{G})$.
However, it is prohibitive to directly optimize the combined risk, since the labeled target samples are indispensable to estimate $\Lambda^{\ell}(\mathbf{C}^*,\mathbf{G})$.

To alleviate the above issue, a simple method  is to use the target pseudo labels with high confidence to estimate the $\Lambda^{\ell}(\mathbf{C}^*,\mathbf{G})$. Given the source samples $\{(\mathbf{x}^i_s, \mathbf{y}^i_s)\}_{i=1}^{n_s}$ and the target samples with high confidence $\{(\mathbf{x}^i_\mathcal{T}, \mathbf{y}^i_\mathcal{T})\}_{i=1}^{n_\mathcal{T}}$, the empirical form of $\Lambda^{\ell}(\mathbf{C}^*,\mathbf{G})$ can be computed by
\begin{equation*}
\frac{1}{n_s}\sum_{i=1}^{n_s} \ell(\mathbf{C}^* \circ \mathbf{G}(\mathbf{x}_s^i), \mathbf{y}_s^i) + \frac{1}{n_\mathcal{T}}\sum_{i=1}^{n_\mathcal{T}} \ell(\mathbf{C}^* \circ \mathbf{G}(\mathbf{x}_{\mathcal{T}}^i), \mathbf{y}_\mathcal{T}^i).
\end{equation*}
However, the combined risk may still increase as shown in Fig \ref{fig:combined_risk} (green line). The reason may be that the target samples, whose pseudo labels with high confidence, are insufficient.

Inspired by semi-supervised learning, an advanced solution is to use \textit{mixup} technique \cite{zhang2018mixup} to augment pseudo-labeled target samples. {Mixup} produces new samples by a convex combination:
given any two samples $(\mathbf{x}_1,\mathbf{y}_1)$, $(\mathbf{x}_2, \mathbf{y}_2)$,
\begin{equation*}
\begin{split}
&\mathbf{x} = \alpha \mathbf{x}_1 + (1-\alpha)\mathbf{x}_2,
~~~\mathbf{y} = \alpha \mathbf{y}_1 + (1-\alpha)\mathbf{y}_2,
\end{split}
\end{equation*}
where $\alpha$ is a hyper-parameter. Zhang et al. \shortcite{zhang2018mixup} has shown that {mixup} not only reduces the memorization to adversarial samples, but also performs better than \textit{Empirical Risk Minimization} (ERM) \cite{vapnik2015uniform}. By applying {mixup} on the target samples with high confidence, new samples $\{(\mathbf{x}^i_m, \mathbf{y}^i_m)\}_{i=1}^n$ are produced, then we propose a proxy $\widetilde{\Lambda}^{\ell}(\mathbf{C}^*, \mathbf{G})$ of $\Lambda^{\ell}$ as follows:
\begin{equation*}
\begin{split}
 \frac{1}{n_s}\sum_{i=1}^{n_s} \ell(\mathbf{C}^* \circ \mathbf{G}(\mathbf{x}_s^i), \mathbf{y}_s^i)+ \frac{1}{n}\sum_{i=1}^n \ell(\mathbf{C}^* \circ \mathbf{G}(\mathbf{x}_m^i), \mathbf{y}_m^i).
\end{split}
\end{equation*}
The aforementioned issue can be mitigated, since {mixup} can be regarded as a data augmentation \cite{zhang2018mixup}. However, the target-domain pseudo labels provided by the source-domain classifier may be inaccurate due to the discrepancy between domains, which causes that mixup may not perform well with inaccurate labels. We propose \textit{enhanced mixup} (e-mixup) to substitute the mixup to compute the proxy.
E-mixup introduces the pure true-labeled source-samples to mitigate the issue caused by bad pseudo labels.

Furthermore, to increase the diversity of new samples, e-mixup produces each new sample using two distant samples, where the distance of the two distant samples is expected to be large. Compared the ordinary mixup technique (i.e., producing new samples using randomly selected samples), e-mixup can produce new examples more effectively. We also verify that e-mixup can further boost the performance (see Table~\ref{tab:aba}). Details of the e-mixup are shown in Algorithm \ref{alg:mixup}.
Corresponding to the double loss situation, denoted by samples $\{(\mathbf{x}^i_e, \mathbf{y}^i_e)\}_{i=1}^n$ produced by e-mixup, the proxy $\widetilde{\Lambda}^{\ell_s\ell_t}(\mathbf{C}^*, \mathbf{G})$ of $\Lambda^{\ell_s \ell_t}$ (defined in Eq. (\ref{DoubleCombinedRisk})) as
\begin{equation}\label{proxycombinedrisk}
\begin{split}
 \frac{1}{n_s}\sum_{i=1}^{n_s} \ell_s(\mathbf{C}^* \circ \mathbf{G}(\mathbf{x}_s^i), \mathbf{y}_s^i)+ \frac{1}{n}\sum_{i=1}^n \ell_t(\mathbf{C}^* \circ \mathbf{G}(\mathbf{x}_e^i), \mathbf{y}_e^i).
\end{split}
\end{equation}
The purple line in Figure~\ref{fig:combined_risk} and ablation study show that e-mixup can further boost performance. 

\subsection{Algorithm}
The optimization of the combined risk plays a crucial role in UDA. Accordingly, we propose a method based on the aforementioned analyses to solve UDA more deeply.
\begin{figure*}[htbp]
\centering
\includegraphics[scale=0.64,trim=0 210 0 170, clip]{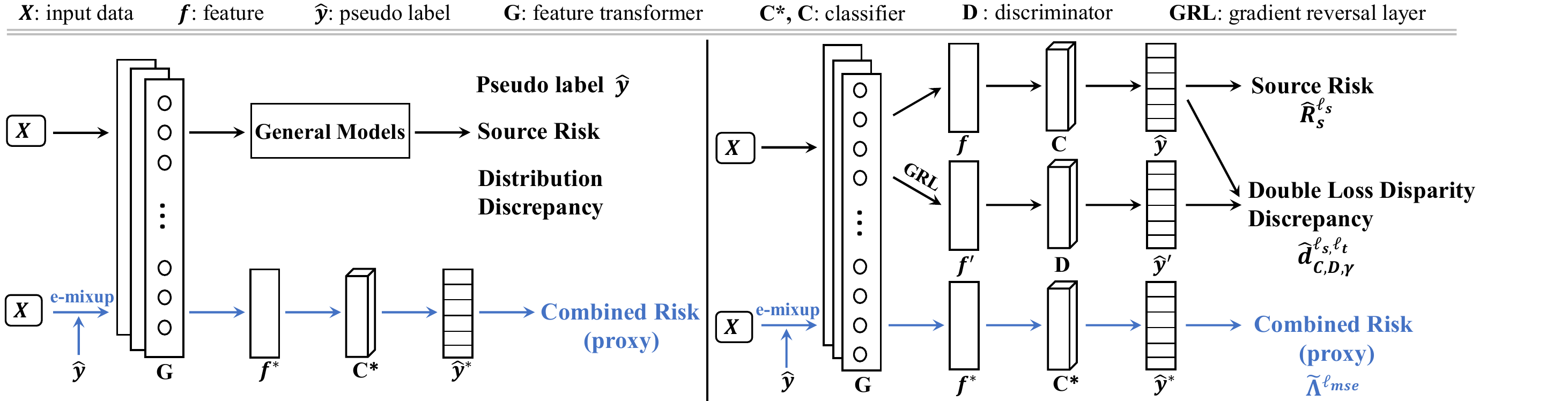}
\caption{The network architecture of applying the proxy of the combined risk. The left figure is the general model for adding the proxy into existing UDA models. The right figure is a specific model based on double loss disparity discrepancy.}
\label{fig:network}
\end{figure*}

\subsubsection{Objective Function}

\begin{algorithm}[t]
	\caption{e-mixup}
	\label{alg:mixup}
	\textbf{Input}: samples $\{(\mathbf{x}^{i}, \mathbf{y}^{i})\}_{i=1}^{n}$.\\
	\textbf{Parameter}: $\alpha$, the number of class $K$. \\
	\textbf{Output}: new samples $\{(\mathbf{x}^{i}_e, \mathbf{y}^{i}_e)\}_{i=1}^{n}$.\\
	\begin{algorithmic}[1]
		\FOR{$i = 1,2,\dots,n$}
        \STATE $y^i = \argmin_{c\in \{1,...,K\}} {y}^i_c$ \% ${y}^i_c$ is the c-th \\coordinate value of vector $\mathbf{y}^i$.
        \STATE Select one from the samples whose label is $y^i$ and denoted it by $(\mathbf{x}^i_r,\mathbf{y}^i_r)$ 
		\STATE $\mathbf{x}_e^i = \alpha \mathbf{x}^i + (1-\alpha)\mathbf{x}^i_r$, 
        ~ $\mathbf{y}_e^i = \alpha \mathbf{y}^i + (1-\alpha)\mathbf{y}^i_r$
        \ENDFOR
	\end{algorithmic}
\end{algorithm}

According to the theoretical bound in Eq. \eqref{eq:bound_dl}, 
we need to solve the
following problem
\begin{equation*}
\begin{split}
\min_{\mathbf{C}\in \mathcal{H},\mathbf{G}\in \mathcal{G}}\big ( \widehat{R}_s^{\ell_s}(\mathbf{C}\circ \mathbf{G})&+d^{\ell_s\ell_t}_{\mathbf{C},\mathcal{H}}(\widehat{P}_{\mathbf{G}(X_s)},\widehat{P}_{\mathbf{G}(X_t)})\\&+\min_{\mathbf{C}^*\in \mathcal{H}}\widetilde{\Lambda}^{\ell_s\ell_t}(\mathbf{C}^*,\mathbf{G}) \big ),
\end{split}
\end{equation*}
where $d^{\ell_s\ell_t}_{\mathbf{C},\mathcal{H}}$ is the double losses disparity discrepancy defined in Eq. \eqref{eq:dbl_dd} and  $\widetilde{\Lambda}^{\ell_s\ell_t}(\mathbf{C}^*,\mathbf{G})$ is defined in Eq. \eqref{proxycombinedrisk}.

Minimizing double loss disparity discrepancy is a \textit{minimax game}, since the double losses disparity discrepancy is defined as the supremum over hypothesis space $\mathcal{H}$. Thus, we revise the above problem as follows:
\begin{equation}\label{FinalOpti2}
\begin{split}
\min_{\mathbf{C}\in \mathcal{H},\mathbf{G}\in \mathcal{G}}  &\big ( \widehat{R}_s^{\ell_s}(\mathbf{C}\circ \mathbf{G})+\widetilde{\Lambda}^{\ell_s\ell_t}(\widetilde{\mathbf{C}},\mathbf{G})\\ +&\eta d_{\mathbf{C},\mathbf{C}^\prime,\gamma}^{\ell_s\ell_t}(\widehat{P}_{\mathbf{G}(X_s)},\widehat{P}_{\mathbf{G}(X_t)}) \big ),
\end{split}
\end{equation}
where $\gamma,\eta$ are parameters to make our model more flexible,
\begin{equation}
    \begin{split}
    &  d_{\mathbf{C},\mathbf{D},\gamma}^{\ell_s\ell_t}(\widehat{P}_{\mathbf{G}(X_s)},\widehat{P}_{\mathbf{G}(X_t)})\\ &=\widehat{R}^{\ell_t}_t(\mathbf{D}\circ \mathbf{G},\mathbf{C} \circ \mathbf{G})-\gamma \widehat{R}^{\ell_s}_s(\mathbf{D}\circ \mathbf{G},\mathbf{C} \circ \mathbf{G}), 
\\ \mathbf{C}^\prime  &= \argmax_{\mathbf{D} \in \mathcal{H}} ~~d_{\mathbf{C},\mathbf{D},\gamma}^{\ell_s\ell_t}(\widehat{P}_{\mathbf{G}(X_s)},\widehat{P}_{\mathbf{G}(X_t)}),
\\  \widetilde{\mathbf{C}}  &= \argmin_{\mathbf{C}^* \in \mathcal{H}}~~\widetilde{\Lambda}^{\ell_s\ell_t}(\mathbf{C}^*,\mathbf{G}).
    \end{split}
\end{equation}


To solve the problem \eqref{FinalOpti2}, we construct a deep method. The network architecture is shown in Fig. \ref{fig:network}(b), which consists of a generator $\mathbf{G}$, a discriminator $\mathbf{D}$, and two classifiers $\mathbf{C}, \mathbf{C}^*$. Next, we introduce the details about our method.


We use standard cross-entropy as the source loss $\ell_s$ and use  modified cross-entropy \cite{NIPS2014_5423, pmlr-v97-zhang19i} as the target loss $\ell_t$. 

For any scoring functions $\mathbf{F},\mathbf{F}^\prime : \mathcal{X}\rightarrow \mathbb{R}^K$,
\begin{equation}
\label{eq:loss}
\begin{split}
&\ell_s(\mathbf{F}(\mathbf{x}),\mathbf{F}^\prime(\mathbf{x})) : = -\log(\sigma_{{{h}^\prime(\mathbf{x})}}(\mathbf{F}(\mathbf{x}))),\\
&\ell_t(\mathbf{F}(\mathbf{x}),\mathbf{F}^\prime(\mathbf{x})) : = \log(1-\sigma_{{{h}^\prime(\mathbf{x})}}(\mathbf{F}(\mathbf{x}))),
\end{split}
\end{equation}
where $\sigma$ is softmax function: for any $\mathbf{y}=[y_1,...,y_K]\in \mathbb{R}^K$,
\begin{equation*}
    \sigma_c(\mathbf{y})=\frac{\exp({{y}_c})}{\sum_{k=1}^K \exp({{y}_k})},~~{\rm for}~c=1,...,K,
\end{equation*}
and
\begin{equation}
    {h}^\prime(\mathbf{x}) = \argmax_{c\in \{1,...,K\}} F^\prime_c(\mathbf{x}),
\end{equation}
here $F^\prime_c$ is the $c$-th coordinate function of function $\mathbf{F}^\prime$.

\textit{Source risk}. Given the source samples $\{(\mathbf{x}^i_s, \mathbf{y}_s^i)\}_{i=1}^{n_s}$, then
\begin{equation}
\label{eq:loss_sr}
\widehat{R}^{\ell_s}_s(\mathbf{C} \circ \mathbf{G}) = \frac{1}{n_s} \sum_{i=1}^{n_s}-\log(\sigma_{{y}_s^i}(\mathbf{C} \circ \mathbf{G}(\mathbf{x}^i_s))),
\end{equation}
where ${y}_s^i$ is the label corresponding to one-hot vector $\mathbf{y}_s^i$.

\textit{Double loss disparity discrepancy.} Given the source and target samples $\{(\mathbf{x}^i_s, \mathbf{y}_s^i)\}_{i=1}^{n_s}, \{\mathbf{x}^i_t\}_{i=1}^{n_t}$, then
\begin{equation}
\label{eq:loss_dd}
\begin{split}
&d^{\ell_s\ell_t}_{\mathbf{C},\mathbf{D},\gamma}(\widehat{P}_{\mathbf{G}(X_s)},\widehat{P}_{\mathbf{G}(X_t)})\\=&-\frac{\gamma}{n_s}\sum_{i=1}^{n_s}\ell_s(\mathbf{D}\circ\mathbf{G}(\mathbf{x}^i_s), \mathbf{C}\circ\mathbf{G}(\mathbf{x}^i_s)) \\&+ \frac{1}{n_t}\sum_{i=1}^{n_t}\ell_t(\mathbf{D}\circ\mathbf{G}(\mathbf{x}^i_t), \mathbf{C}\circ \mathbf{G}(\mathbf{x}^i_t)),
\end{split}
\end{equation}
where $\ell_s, \ell_t$ are defined in Eq. \eqref{eq:loss}.

\textit{Combined risk}. As discussed in Motivation, the combined risk cannot  be optimized directly. To mitigate this problem, we use the proxy $\widetilde{\Lambda}^{\ell_s\ell_t}$ in Eq. \eqref{proxycombinedrisk} to substitute it.

Further, motivated by \cite{berthelot2019mixmatch}, we use mean mquare error ($\ell_{mse}$) to calculate the proxy of the combined risk, because, unlike the cross-entropy loss, $\ell_{mse}$ is bounded and less sensitive to incorrect predictions.
Denoted by $\{(\mathbf{x}^i_{e}, \mathbf{y}^i_{e})\}_{i=1}^{n}$ the output of the \textit{e-mixup}. Then the proxy is calculated by
\begin{equation}\label{eq:loss_lbd}
\begin{split}
\widetilde{\Lambda}^{\ell_{mse}}(\mathbf{C}^*, \mathbf{G}) =& \frac{1}{n_s}\sum_{i=1}^{n_s} \ell_{mse}(\mathbf{C}^* \circ \mathbf{G}(\mathbf{x}_s^i), \mathbf{y}_s^i) \\&+ \frac{1}{n}\sum_{i=1}^n \ell_{mse}(\mathbf{C}^* \circ \mathbf{G}(\mathbf{x}_e^i), \mathbf{y}_e^i).
\end{split}
\end{equation}

\begin{algorithm}[ht]
	\caption{The training procedure of E-MixNet}
	\label{alg:training}
	\textbf{Input}: source, target samples $\{(\mathbf{x}^{i}_s, \mathbf{y}^{i}_s)\}_{i=1}^{n_s}$,$\{\mathbf{x}^{i}_{t}\}_{i=1}^{n_t}$.\\
	\textbf{Parameter}: learning rate $l$, batch size $n_b$, the number of iteration $T$, network parameters $\mathbf{\Theta}$.\\
	\textbf{Output}: the predicted target label $\widehat{\mathbf{y}}_t$.\\
	\begin{algorithmic}[1]
		\STATE Initialize $\mathbf{\Theta}$
		\FOR{$j = 1,2,\dots,T$}
		\STATE Fetch source minibatch $D_s^m$
		\STATE Fetch target minibatch $D_t^m$ 
	    \STATE Calculate $\widehat{R}^{\ell_s}_s$ using $D_s^m$ (Eq. (\ref{eq:loss_sr}))
		\STATE Calculate $d^{\ell_s\ell_t}_{\mathbf{C},\mathbf{D},\gamma}$ using $D_s^m, D_t^m$ (Eq. \eqref{eq:loss_dd})
		\STATE Obtain highly confident target samples $\{(\mathbf{x}^{i}_\mathcal{T}, \mathbf{y}^{i}_\mathcal{T})\}_{i=1}^{n_\mathcal{T}}$ predicted by $\mathbf{C}\circ \mathbf{G}$ on $D_t^m$
        \STATE $\{(\mathbf{x}^i, \mathbf{y}^i)\}_{i=1}^n = D_s^m \cup \{(\mathbf{x}^{i}_\mathcal{T}, \mathbf{y}^{i}_\mathcal{T})\}_{i=1}^{n_\mathcal{T}}$
		\STATE $\{(\mathbf{x}^i_e, \mathbf{y}^i_e)\}_{i=1}^n$ = e-mixup($\{(\mathbf{x}^i, \mathbf{y}^i)\}_{i=1}^n$)
		\STATE Calculate $\widetilde{\Lambda}^{\ell_{mse}}$ using $D_s^m,\{(\mathbf{x}^i_e, \mathbf{y}^i_e)\}_{i=1}^n$  (Eq. \eqref{eq:loss_lbd})
		\STATE Update $\mathbf{\Theta}$ according to Eq. \eqref{FinalOpti3}
		\ENDFOR
	\end{algorithmic}
\end{algorithm}

\begin{table*}[htbp]
\caption{Results on Office-31 (ResNet-50)}
\label{tab:of31}
\vspace{-1em}
\centering
\begin{tabular}{l|ccccccc} \hline
Method & A$\rightarrow$W &D$\rightarrow$W &W$\rightarrow$D &A$\rightarrow$D &D$\rightarrow$A &W$\rightarrow$A &Avg \\ \hline
ResNet-50 \cite{he2016deep}              &68.4$\pm$0.2 &96.7$\pm$0.1 &99.3$\pm$0.1 &68.9$\pm$0.2 &62.5$\pm$0.3 &60.7$\pm$0.3 &76.1\\
DAN \cite{long2015learning} 		        &80.5$\pm$0.4 &97.1$\pm$0.2 &99.6$\pm$0.1 &78.6$\pm$0.2 &63.6$\pm$0.3 &62.8$\pm$0.2 &80.4\\
RTN \cite{long2016unsupervised} 		    &84.5$\pm$0.2 &96.8$\pm$0.1 &99.4$\pm$0.1 &77.5$\pm$0.3 &66.2$\pm$0.2 &64.8$\pm$0.3 &81.6\\
DANN \cite{ganin2016domain} 		        &82.0$\pm$0.4 &96.9$\pm$0.2 &99.1$\pm$0.1 &79.7$\pm$0.4 &68.2$\pm$0.4 &67.4$\pm$0.5 &82.2\\
ADDA \cite{tzeng2017adversarial} 		&86.2$\pm$0.5 &96.2$\pm$0.3 &98.4$\pm$0.3 &77.8$\pm$0.3 &69.5$\pm$0.4 &68.9$\pm$0.5 &82.9\\
JAN \cite{long2013transfer} 		        &86.0$\pm$0.4 &96.7$\pm$0.3 &99.7$\pm$0.1 &85.1$\pm$0.4 &69.2$\pm$0.3 &70.7$\pm$0.5 &84.6\\
MADA \cite{pei2018multi} 		        &90.0$\pm$0.1 &97.4$\pm$0.1 &99.6$\pm$0.1 &87.8$\pm$0.2 &70.3$\pm$0.3 &66.4$\pm$0.3 &85.2\\
SimNet \cite{pinheiro2018unsupervised}   &88.6$\pm$0.5 &98.2$\pm$0.2 &99.7$\pm$0.2 &85.3$\pm$0.3 &73.4$\pm$0.8 &71.8$\pm$0.6 &86.2\\
MCD \cite{saito2018maximum} 		        &89.6$\pm$0.2 &98.5$\pm$0.1 &\textbf{100.0}$\pm$.0 &91.3$\pm$0.2 &69.6$\pm$0.1 &70.8$\pm$0.3 &86.6\\
CDAN+E \cite{long2018conditional} 		&94.1$\pm$0.1 &98.6$\pm$0.1 &\textbf{100.0}$\pm$.0 &92.9$\pm$0.2 &71.0$\pm$0.3 &69.3$\pm$0.3 &87.7\\
SymNets \cite{zhang2019domain}           &90.8$\pm$0.1 &98.8$\pm$0.3 &\textbf{100.0}$\pm$.0 &93.9$\pm$0.5 &74.6$\pm$0.6 &72.5$\pm$0.5 &88.4\\
MDD \cite{pmlr-v97-zhang19i} 		    &\textbf{94.5}$\pm$0.3 &98.4$\pm$0.1 &\textbf{100.0}$\pm$.0 &93.5$\pm$0.2 &74.6$\pm$0.3 &72.2$\pm$0.1 &88.9\\
{E-MixNet}                              &93.0$\pm$0.3 &\textbf{99.0}$\pm$0.1 &\textbf{100.0}$\pm$.0 &\textbf{95.6}$\pm$0.2 &\textbf{78.9}$\pm$0.5 &\textbf{74.7}$\pm$0.7 &\textbf{90.2} \\ 
\hline
\end{tabular}
\vspace{-0.5em}
\end{table*}

\begin{table*}[htbp]
\caption{Results on Image-CLEF (ResNet-50)}
\label{tab:image}
\vspace{-1em}
\centering
\begin{tabular}{l|ccccccc} \hline
Method & I$\rightarrow$P &P$\rightarrow$I &I$\rightarrow$C &C$\rightarrow$I &C$\rightarrow$P &P$\rightarrow$C &Avg \\ \hline
ResNet-50 \cite{he2016deep}      &74.8$\pm$0.3 &83.9$\pm$0.1 &91.5$\pm$0.3 &78.0$\pm$0.2 &65.5$\pm$0.3 &91.2$\pm$0.3 &80.7\\
DAN \cite{long2015learning}  	&74.5$\pm$0.4 &82.2$\pm$0.2 &92.8$\pm$0.2 &86.3$\pm$0.4 &69.2$\pm$0.4 &89.8$\pm$0.4 &82.5\\
DANN \cite{ganin2016domain} 		&75.0$\pm$0.6 &86.0$\pm$0.3 &96.2$\pm$0.4 &87.0$\pm$0.5 &74.3$\pm$0.5 &91.5$\pm$0.6 &85.0\\
JAN \cite{long2013transfer} 		&76.8$\pm$0.4 &88.0$\pm$0.2 &94.7$\pm$0.2 &89.5$\pm$0.3 &74.2$\pm$0.3 &91.7$\pm$0.3 &85.8\\
MADA \cite{pei2018multi} 		&75.0$\pm$0.3 &87.9$\pm$0.2 &96.0$\pm$0.3 &88.8$\pm$0.3 &75.2$\pm$0.2 &92.2$\pm$0.3 &85.8\\
CDAN+E \cite{long2018conditional}&77.7$\pm$0.3 &90.7$\pm$0.2 &\textbf{97.7}$\pm$0.3 &91.3$\pm$0.3 &74.2$\pm$0.2 &94.3$\pm$0.3 &87.7\\
SymNets \cite{zhang2019domain}   &80.2$\pm$0.3 &93.6$\pm$0.2 &97.0$\pm$0.3 &93.4$\pm$0.3 &78.7$\pm$0.3 &96.4$\pm$0.1 &89.9\\
E-MixNet  	                    &\textbf{80.5}$\pm$0.4 &\textbf{96.0}$\pm$0.1 &\textbf{97.7}$\pm$0.3 &\textbf{95.2}$\pm$0.4 &\textbf{79.9}$\pm$0.2 &\textbf{97.0}$\pm$0.3 &\textbf{91.0} \\
\hline
\end{tabular}
\vspace{-1em}
\end{table*}

\subsubsection{Training Procedure}
Finally, the UDA problem can be solved by the following minimax game.
\begin{equation}\label{FinalOpti3}
\begin{split}
&\min_{\mathbf{C}\in \mathcal{H},\mathbf{G}\in \mathcal{G}} \big ( \widehat{R}_s^{\ell_s}(\mathbf{C}\circ \mathbf{G})+\widetilde{\Lambda}^{\ell_{ mse}}({\mathbf{C}}^*,\mathbf{G})\\&~~~~~~~~~~~~~~~~~+\eta d_{\mathbf{C},\mathbf{D},\gamma}^{\ell_s\ell_t}(\widehat{P}_{\mathbf{G}(X_s)},\widehat{P}_{\mathbf{G}(X_t)}) \big ),
\\&~~~~ \max_{\mathbf{D} \in \mathcal{H}} ~~~d_{\mathbf{C},\mathbf{D},\gamma}^{\ell_s\ell_t}(\widehat{P}_{\mathbf{G}(X_s)},\widehat{P}_{\mathbf{G}(X_t)}),
\\&~~~~\min_{\mathbf{C}^* \in \mathcal{H}}~~~\widetilde{\Lambda}^{\ell_{ mse}}(\mathbf{C}^*,\mathbf{G}).
\end{split}
\end{equation}
The training procedure is shown in Algorithm 2.

\section{Experiments}

\begin{table*}[ht]
\centering
\caption{Results on Office-Home (ResNet-50)}
\vspace{-0.8em}
\label{tab:OfHome}
\begin{tabular}{l|p{0.6cm}<{\centering}p{0.6cm}<{\centering}p{0.6cm}<{\centering}p{0.6cm}<{\centering}p{0.6cm}<{\centering}p{0.6cm}<{\centering}p{0.6cm}<{\centering}p{0.6cm}<{\centering}p{0.6cm}<{\centering}p{0.6cm}<{\centering}p{0.6cm}<{\centering}p{0.6cm}<{\centering}p{0.6cm}<{\centering}} \hline
Method &A$\rightarrow$C &A$\rightarrow$P &A$\rightarrow$R &C$\rightarrow$A &C$\rightarrow$P &C$\rightarrow$R &P$\rightarrow$A &P$\rightarrow$C &P$\rightarrow$R &R$\rightarrow$A &R$\rightarrow$C &R$\rightarrow$P &Avg\\ \hline
ResNet-50 \cite{he2016deep}          &~34.9 &~50.0 &~58.0 &~37.4 &~41.9 &~46.2 &~38.5 &~31.2 &~60.4 &~53.9 &~41.2 &~59.9 &46.1\\
DAN \cite{long2015learning}  		&~54.0 &~68.6 &~75.9 &~56.4 &~66.0 &~67.9 &~57.1 &~50.3 &~74.7 &~68.8 &~55.8 &~80.6 &64.7\\
DANN \cite{ganin2016domain}		    &~44.1 &~66.5 &~74.6 &~57.9 &~62.0 &~67.2 &~55.7 &~40.9 &~73.5 &~67.5 &~47.9 &~77.7 &61.3\\
JAN \cite{long2013transfer} 		    &~45.9 &~61.2 &~68.9 &~50.4 &~59.7 &~61.0 &~45.8 &~43.4 &~70.3 &~63.9 &~52.4 &~76.8 &58.3\\
CDAN+E \cite{long2018conditional} 	&~47.0 &~69.4 &~75.8 &~61.0 &~68.8 &~70.8 &~60.2 &~47.1 &~77.9 &~70.8 &~51.4 &~81.7 &65.2\\
SymNets \cite{zhang2019domain}  	    &~46.0 &~73.8 &~78.2 &~\textbf{64.1} &~69.7 &~74.2 &~63.2 &~48.9 &~\textbf{80.0} &~\textbf{74.0} &~51.6 &~82.9 &67.2\\
MDD \cite{pmlr-v97-zhang19i}  		&~54.9 &~73.7 &~77.8 &~60.0 &~71.4 &~71.8 &~61.2 &~53.6 &~78.1 &~72.5 &~60.2 &~82.3 &68.1\\
{E-MixNet}                          &~\textbf{57.7} &~\textbf{76.6} &~\textbf{79.8} &~63.6 &~\textbf{74.1} &~\textbf{75.0} &~\textbf{63.4} &~\textbf{56.4} &~79.7 &~72.8 &~\textbf{62.4} &~\textbf{85.5} &\textbf{70.6} \\ 
\hline
\end{tabular}
\vspace{-0.5em}
\end{table*}

\begin{table*}[htbp]
\centering
\caption{The results of combination experiments on Office-Home (ResNet-50)}
\vspace{-1em}
\label{tab:OfHome_lbd}
\begin{tabular}{l|p{0.6cm}<{\centering}p{0.6cm}<{\centering}p{0.6cm}<{\centering}p{0.6cm}<{\centering}p{0.6cm}<{\centering}p{0.6cm}<{\centering}p{0.6cm}<{\centering}p{0.6cm}<{\centering}p{0.6cm}<{\centering}p{0.6cm}<{\centering}p{0.6cm}<{\centering}p{0.6cm}<{\centering}p{0.6cm}<{\centering}} \hline
Method &A$\rightarrow$C &A$\rightarrow$P &A$\rightarrow$R &C$\rightarrow$A &C$\rightarrow$P &C$\rightarrow$R &P$\rightarrow$A &P$\rightarrow$C &P$\rightarrow$R &R$\rightarrow$A &R$\rightarrow$C &R$\rightarrow$P &Avg\\ \hline
DAN \cite{long2015learning}                     &~54.0 &~68.6 &~75.9 &~56.4 &~66.0 &~67.9 &~57.1 &~50.3 &~74.7 &~68.8 &~55.8 &~80.6 &64.7 \\
DAN+$\widetilde{\Lambda}^{\ell_{mse}}$          &~\textbf{57.0} &~\textbf{71.0} &~\textbf{77.9} &~\textbf{59.9} &~\textbf{72.6} &~\textbf{70.1} &~\textbf{58.1} &~\textbf{57.1} &~\textbf{77.3} &~\textbf{72.7} &~\textbf{64.7} &~\textbf{84.6} &\textbf{68.6} \\\hline
DANN \cite{ganin2016domain}                     &~44.1 &~66.5 &~74.6 &~57.9 &~62.0 &~67.2 &~55.7 &~40.9 &~73.5 &~67.5 &~47.9 &~77.7 &61.3 \\
DANN+$\widetilde{\Lambda}^{\ell_{mse}}$         &~\textbf{50.9} &~\textbf{69.6} &~\textbf{77.8} &~\textbf{61.9} &~\textbf{70.7} &~\textbf{71.6} &~\textbf{60.0} &~\textbf{49.5} &~\textbf{78.4} &~\textbf{71.8} &~\textbf{55.7} &~\textbf{83.7} &\textbf{66.8}\\ \hline
CDAN+E \cite{long2018conditional}               &~47.0 &~69.4 &~75.8 &~61.0 &~68.8 &~70.8 &~60.2 &~47.1 &~77.9 &~70.8 &~51.4 &~81.7 &65.2 \\ 
CDAN+E+$\widetilde{\Lambda}^{\ell_{mse}}$       &~\textbf{49.5} &~\textbf{70.1} &~\textbf{77.8} &~\textbf{64.3} &~\textbf{71.3} &~\textbf{74.2} &~\textbf{61.6} &~\textbf{50.6} &~\textbf{80.0} &~\textbf{73.5} &~\textbf{56.6} &~\textbf{84.1} &\textbf{67.8}\\ \hline
SymNets \cite{zhang2019domain}                  &~46.0 &~73.8 &~78.2 &~64.1 &~69.7 &~74.2 &~63.2 &~\textbf{48.9} &~80.0 &~74.0 &~51.6 &~82.9 &67.2\\
SymNets+$\widetilde{\Lambda}^{\ell_{mse}}$      &~\textbf{48.8} &~\textbf{74.7} &~\textbf{79.7} &~\textbf{64.9} &~\textbf{72.5} &~\textbf{75.6} &~\textbf{63.9} &~47.0 &~\textbf{80.8} &~\textbf{73.9} &~\textbf{52.4} &~\textbf{83.9} &\textbf{68.2} \\ 
\hline
\end{tabular}
\vspace{-1em}
\end{table*}


\begin{table}[htbp]
\caption{Ablation experiments on Image-CLEF}
\label{tab:aba}
\vspace{-0.8em}
\centering
\begin{tabular}{p{0.08cm}p{0.08cm}p{0.08cm}p{0.08cm}|p{0.45cm}p{0.45cm}p{0.45cm}p{0.45cm}p{0.45cm}p{0.45cm}p{0.45cm}} \hline
s &t &m &e
&I$\rightarrow$P &P$\rightarrow$I &I$\rightarrow$C &C$\rightarrow$I &C$\rightarrow$P &P$\rightarrow$C &Avg\\ \hline
&&&                                                 &80.2 &94.2 &96.7 &94.7 &79.2 &95.5 &90.1\\
&$\small{\surd}$&&                                  &79.9 &92.2 &97.7 &93.8 &79.4 &96.5 &89.9\\
&$\small{\surd}$&$\small{\surd}$&                   &79.7 &93.7 &97.5 &94.5 &79.7 &96.2 &90.2\\
$\small{\surd}$&$\small{\surd}$&$\small{\surd}$&    &79.4 &95.0 &\textbf{97.8} &94.8 &\textbf{81.4} &96.5 &90.8\\
$\small{\surd}$&$\small{\surd}$& &$\small{\surd}$   &\textbf{80.5} &\textbf{96.0} &97.7 &\textbf{95.2} &79.9 &\textbf{97.0} &\textbf{91.0}\\	\hline
\end{tabular}
\vspace{-1em}
\end{table}

We evaluate E-Mixnet on three public datasets, and compare it with several existing state-of-the-art methods. Codes will be available at https://github.com/zhonglii/E-MixNet.

\subsection{Datasets}
Three common UDA datasets are used to evaluate the efficacy of E-MixNet. 

\textbf{Office-31} \cite{saenko2010adapting} is an object recognition dataset with $4,110$ images, which consists of three domains with a slight discrepancy: \textit{amazon} ({A}), \textit{dslr} ({D}) and \textit{webcam} ({W}). Each domain contains $31$ kinds of objects. So there are $6$ domain adaptation tasks on \textit{Office-31}: {A} $\rightarrow$ {D}, {A} $\rightarrow$ {W}, {D} $\rightarrow$ {A}, {D} $\rightarrow$ {W}, {W} $\rightarrow$ {A}, {W} $\rightarrow$ {D}. 

\textbf{Office-Home} \cite{venkateswara2017deep} is an object recognition dataset with $15,500$ image, which contains four domains with more obvious domain discrepancy than \textit{Office-31}. These domains are \textit{Artistic} ({A}), \textit{Clipart} ({C}), \textit{Product} ({P}), \textit{Real-World} ({R}). Each domain contains $65$ kinds of objects. So there are $12$ domain adaptation tasks on {Office-Home}: {A} $\rightarrow$ {C}, {A} $\rightarrow$ {P}, {A} $\rightarrow$ {R}, ..., {R} $\rightarrow$ {P}. 

\textbf{ImageCLEF-DA}\footnote{http://imageclef.org/2014/adaptation/} is a dataset organized by selecting the 12 common classes shared by three public datasets (domains): \textit{Caltech-256} (C), \textit{ImageNet ILSVRC} 2012 (I), and \textit{Pascal VOC} 2012 (P). We permute all three domains and build six transfer tasks: I$\rightarrow$P, P$\rightarrow$I, I$\rightarrow$C, C$\rightarrow$I, C$\rightarrow$P, P$\rightarrow$C.

\subsection{Experimental Setup}
Following the standard protocol for unsupervised domain adaptation in \cite{ganin2016domain, long2018conditional}, all labeled source  samples and unlabeled target samples are used in the training process and we report the average classification accuracy based on three random experiments. $\gamma$ in Eq. \eqref{eq:loss_dd} is selected from {2, 4, 8}, and it is set to 2 for Office-Home, 4 for Office-31, and 8 for Image-CLEF.

ResNet-50 \cite{he2016deep} pretrained on ImageNet is employed as the backbone network ($\mathbf{G}$). $\mathbf{C}$, $\mathbf{D}$ and $\mathbf{C}^*$ are all two fully connected layers where the hidden unit is 1024. Gradient reversal layer between \textbf{G} and $\mathbf{D}$ is employed for adversarial training. The algorithm is implemented by Pytorch. The mini-batch stochastic gradient descent with momentum 0.9 is employed as the optimizer, and the learning rate is adjected by $l_i = l_0(1+\delta i)^{-\beta}$, where i linearly increase from 0 to 1 during the training process, $\delta=10$, $l_0=0.04$. We follow \cite{pmlr-v97-zhang19i} to employ a progressive strategy for $\eta$: $\eta=\frac{2\eta_0}{1+\exp(\delta*i)-\eta_0}$, $\eta_0$ is set to 0.1. The $\alpha$ in e-mixup is set to 0.6 in all experiments.

\subsection{Results}
\vspace{-0.125em}
The results on \textit{Office-31} are reported in Tabel \ref{tab:of31}. E-MixNet achieves the best results and exceeds the baselines for 4 of 6 tasks. Compared to the competitive baseline MDD, E-MixNet surpasses it by 4.3\% for the difficult task D $\rightarrow$ A.

The results on \textit{Image-CLEF} are reported in Table \ref{tab:image}. E-MixNet significantly outperforms the baselines for 5 of 6 tasks. For the hard task C $\rightarrow$ P, E-MixNet surpasses the competitive baseline SymNets by 2.7\%.

The results on \textit{Office-Home} are reported in Table \ref{tab:OfHome}. Despite \textit{Office-Home} is a challenging dataset, E-MixNet still achieves better performance than all the baselines for 9 of 12 tasks. For the difficult tasks A $\rightarrow$ C, P $\rightarrow$ A, and R $\rightarrow$ C, E-MixNet has significant advantages.

In order to further verify the efficacy of the proposed proxy of the combined risk, we add the proxy into the loss functions of four representative UDA methods. As shown in Fig. \ref{fig:network}(a), we add a new classifier that is the same as the classifier in the original method to formulate the proxy of the combined risk. The results are shown in Table \ref{tab:OfHome_lbd}. The four methods can achieve better performance after optimizing the proxy. It is worth noting that DANN obtains a 5.5\% percent increase. The experiments adequately demonstrate the combined risk plays a crucial role for methods that aim to learn a domain-invariant representation and the proxy can indeed curb the increase of the combined risk.

\subsection{Ablation Study and Parameter Analysis}
\textit{Ablation Study}. To further verify the efficacy of the proxy of combined risk calculated by mixup and e-mixup respectively. Ablation experiments are shown in Tabel \ref{tab:aba}, where s indicates that the source samples are introduced to augment the target samples, t indicates augmenting the target samples, m denotes mixup, and e denotes e-mixup. Table \ref{tab:aba} shows that E-MixNet achieves the best performance, which further shows that we can effectively control the combined risk by the proxy $\widetilde{\Lambda}^{{\ell}_{mse}}$.

\begin{figure}[!t]
\centering
\includegraphics[scale=0.3,trim=5 16 0 0, clip]{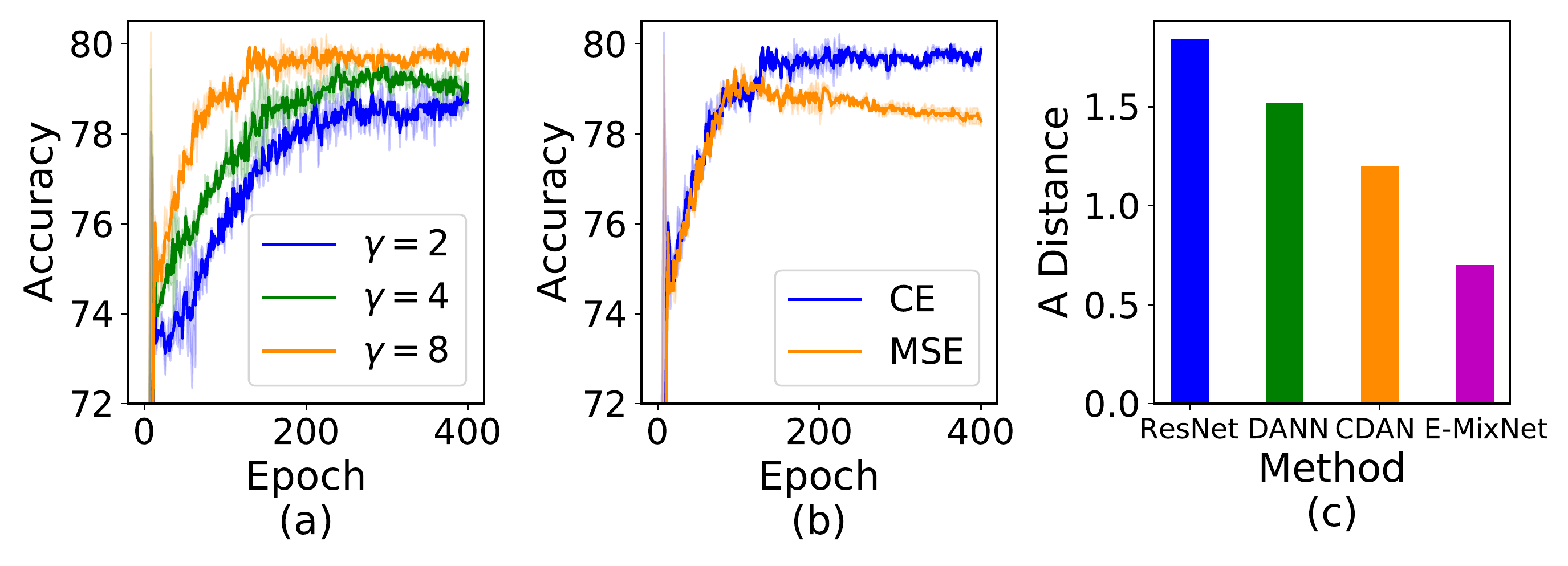}
\caption{The impact of $\gamma$ is shown in (a). The impact of the loss functions for the proxy of the combined risk is shown in (b). Comparison of $\mathcal{A}$-distance.
    }
\label{fig:param}
\vspace{-1.5em}
\end{figure}

\noindent \textit{Parameter analysis}. 
Here we aim to study how the parameter $\gamma$ affects the performance and the efficiencies of \emph{mean square error} (MSE) and cross-entropy for the proxy of combined risk. Firstly, as shown in Fig. \ref{fig:param}(a), a relatively larger $\gamma$ can obtain better performance and faster convergence. Secondly, when \textit{mixup} behaves between two samples, the accuracy of the pseudo labels of the target samples are much important. To against the adversarial samples, MSE is employed to substitute cross-entropy. As shown in Fig. \ref{fig:param}(b), MSE can obtain more stable and better performance. 
Furthermore, $\mathcal{A}$-distance is also an important indicator showing the distribution discrepancy, which is defined as $dis_{\mathcal{A}}=2(1-2\epsilon)$ where $\epsilon$ is the test error. As shown in Fig. \ref{fig:param} (c). E-MixNet achieves a better performance of adaptation, implying the efficiency of the proposed proxy.

\vspace{-0.2cm}
\section{Conclusion}
Though numerous UDA methods have been proposed and have achieved significant success, the issue caused by combined risk has not been brought to the forefront and none of the proposed methods solve the problem.
Theorem 3 reveals that the combined risk is deeply related to the conditional distribution discrepancy and plays a crucial role for transfer performance.
Furthermore, we propose a method termed E-MixNet, which employs \textit{enhanced mixup} to calculate a proxy of the combined risk.
Experiments show that our method achieves a comparable performance compared with existing state-of-the-art methods and the performance of the four representative methods can be boosted by adding the proxy into their loss functions.
\section{Acknowledgments}
The  work  presented  in  this  paper  was  supported  by  the Australian Research Council (ARC) under DP170101632 and FL190100149. The first author particularly thanks the support of UTS-AAII during his visit.


\bibliography{main.bbl}
\end{document}